\documentclass[]{elsarticle}

\usepackage{lineno,hyperref}
\usepackage{microtype}

\usepackage{graphicx}
\usepackage{amsmath}
\usepackage{multirow}
\usepackage{booktabs}
\usepackage{amsfonts}
\usepackage{bbm}
\usepackage{float}
\usepackage{subfig}
\usepackage{algorithmic}
\usepackage{caption}

\usepackage[english]{babel}

\usepackage{bm}

\modulolinenumbers[5] 

\journal{Journal of \LaTeX\ Templates}

\usepackage{times,latexsym}
\usepackage{url}
\usepackage[T1]{fontenc}
\usepackage{amsfonts}
\usepackage{amsthm}
\usepackage{amssymb}
\usepackage{booktabs}
\usepackage{multirow}
\usepackage{xcolor}
\usepackage{hyperref}

\usepackage[linesnumbered,vlined,ruled]{algorithm2e}
\usepackage{tikz}
\usepackage{tkz-graph}
\SetVertexNormal[Shape = circle, FillColor  = orange,LineWidth  = 2pt]
\SetUpEdge[lw = 1.5pt,color = black, labelcolor = white, labeltext  = red,
labelstyle = {sloped,draw,text=blue}]

\usetikzlibrary{arrows, decorations.markings}

\tikzstyle{vecArrow} = [thick, decoration={markings,mark=at position
	1 with {\arrow[semithick]{open triangle 60}}},
double distance=2pt, shorten >= 5.5pt,
preaction = {decorate},
postaction = {draw,line width=2pt, white,shorten >= 4.5pt}]
\tikzstyle{innerWhite} = [semithick, white,line width=2pt, shorten >= 4.5pt]

\newtheorem{proposition}{Proposition}

\makeatletter
\DeclareRobustCommand{\sqcdot}{\mathbin{\mathpalette\morphic@sqcdot\relax}}
\newcommand{\morphic@sqcdot}[2]{%
	\sbox\z@{$\m@th#1\centerdot$}%
	\ht\z@=.33333\ht\z@
	\vcenter{\box\z@}%
}
\begin{document}

\begin{frontmatter}

\title{Boolean Product Graph Neural Networks}

\author[address1]{Ziyan Wang}



\author[address1]{Bin Liu\corref{equ}\corref{corresponding author}}
\cortext[corresponding author]{Corresponding author}
\ead{liubin@swufe.edu.cn}
\author[address1]{Ling Xiang}

\address[address1]{The Center of Statistical Research, School of Statistics, \\ Southwestern University of Finance and Economics, Chengdu, China }





\begin{abstract}
Graph Neural Networks (GNNs) have recently achieved significant success, with a key operation involving the aggregation of information from neighboring nodes. Substantial researchers have focused on defining neighbors for aggregation, predominantly based on observed adjacency matrices. However, in many scenarios, the explicitly given graphs contain noise, which can be amplified during the messages-passing process. Therefore, many researchers have turned their attention to latent graph inference, specifically learning a parametric graph. To mitigate fluctuations in latent graph structure learning, this paper proposes a novel Boolean product-based graph residual connection in GNNs to link the latent graph and the original graph. It computes the Boolean product between the latent graph and the original graph at each layer to correct the learning process. The Boolean product between two adjacency matrices is equivalent to triangle detection. 
Accordingly, the proposed Boolean product graph neural networks can be interpreted as discovering triangular cliques from the original and the latent graph. We validate the proposed method in benchmark datasets and demonstrate its ability to enhance the performance and robustness of GNNs.
\end{abstract}

\begin{keyword} Boolean Product\sep Graph Neural Networks\sep
Latent Graph Inference
\end{keyword}


\end{frontmatter}

\newpage
\section{Introduction}
In recent years, graph neural networks (GNNs) \cite{defferrard2016convolutional,kipf2017semi,hamilton2017inductive,veli2018graph} have achieved significant success in the analysis of graph data. 
By aggregating information from multiple hops and integrating the feature extraction capability of traditional deep learning, GNNs have been successfully applied to various graph data analysis tasks,
such as microscopic molecular networks \cite{li2018adaptive}, protein networks \cite{strokach2020fast}, as well as macroscopic social networks, traffic networks \cite{wang2020traffic}, and industrial chain \cite{ijcai2023p674}.
In the aforementioned scenarios, the graph is given or observed, and traditional GNNs essentially adhere to this assumption. 

However, the assumption of a given graph restricts the ability of GNNs. In some scenarios, there exist significant relationships between samples, but there is no directly observable graph available~\cite{franceschi2019learning,wang2019dynamic}. Even though the graph is given~\cite{topping2022understanding,chen2020iterative,sun2023self}, there are observational errors that introduce graph noise in some cases~\cite{franceschi2019learning}, which is gradually amplified by the aggregation of neighbor information and severely impact the performance.
For example, in studies on the toxicity of compounds, it is challenging to regress the toxicity of compounds based on molecular graphs~\cite{li2018adaptive}. 
Similarly, in the domain of macroscopic social networks, connections among individuals on social media may not effectively reflect their underlying relevance.

\begin{figure*}[!htp]
  \centering
   \includegraphics[width=1.0\linewidth]{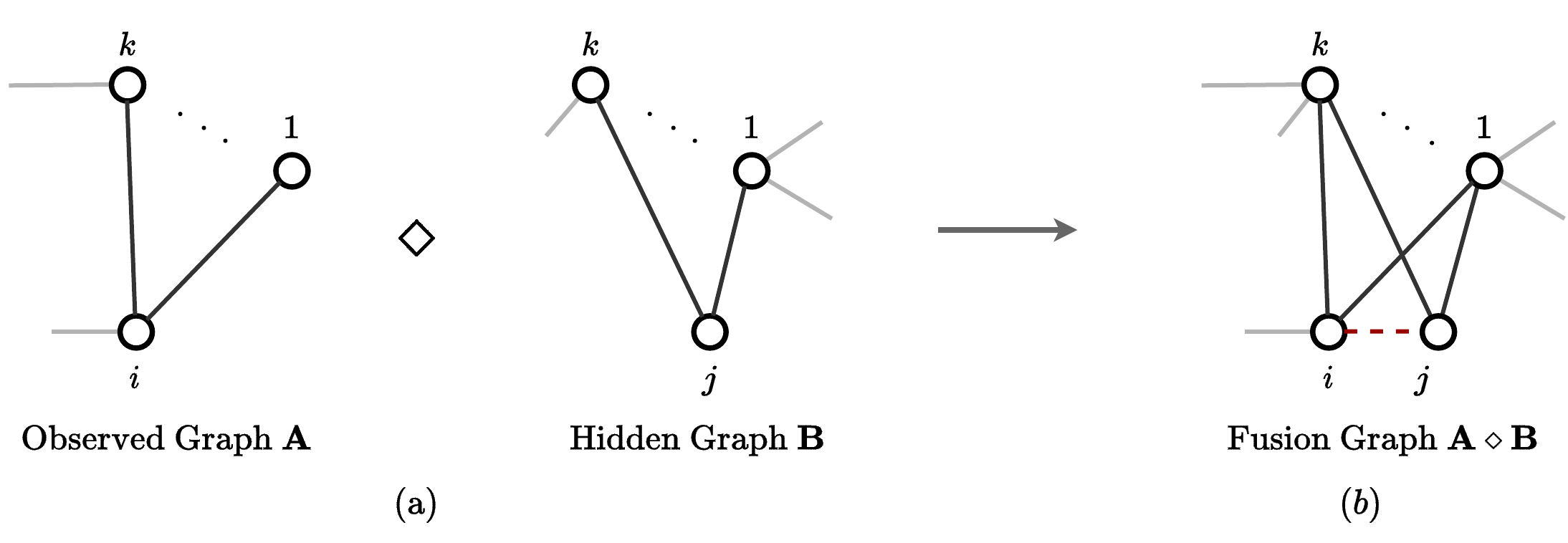}
   \caption{ The interpretation of Boolean product between two adjacent matrices $\mathbf{A}$ and $\mathbf{B}$. The graphs defined on $\mathbf{A}$ and $\mathbf{B}$ share a same set of nodes, and all are aligned.
   Suppose that there is no edge between node $i$ and $j$ from the perspectives of both matrices $\mathbf{A}$ and $\mathbf{B}$, that is $\mathbf{A}_{ij}=0$ and $\mathbf{B}_{ij}=0$. Panel (a) demonstrates the set of shared neighbors $\mathcal{N}_i(\mathbf{A}) \cap \mathcal{N}_j(\mathbf{B}):=\{1,\cdots,k\}$ of node $i$ from view $\mathbf{A}$ and node $j$ from view $\mathbf{B}$. Panel (b) visualizes the Boolean product resulting matrix $\mathbf{A}\diamond \mathbf{B}$ (as shown in \autoref{eq:booleanFusion}). We observe $[\mathbf{A}\diamond \mathbf{B}]_{ij}=1$ once we observe that the shared neighbor set $\{1,\cdots,k\}$ of nodes $i$ and $j$ is not null. It yields a new edge (red dash line) between node $i$ and $j$ as the overall relationship from the points of view of both $\mathbf{A}$ and $\mathbf{B}$.
   } 
   \label{fig:BooleanProduct}
\end{figure*}

Researchers recently turn their focus towards latent graph inference~\cite{fatemi2021slaps,kazi2022differentiable,de2022Latent} to address issues of the absence of an observational graph or encountering noise in given graphs. In situations where the graph is unknown, attention mechanisms prove valuable for inferring dynamic relationships among samples~\cite{franceschi2019learning,hu2023futures}. Conversely, when a graph is provided, latent graph inference becomes crucial for improving predictive performance by rectifying the process of aggregating neighbor information~\cite{topping2022understanding,kazi2022differentiable}.
However, unlike fixed graphs, latent graphs are parameterized and need to be updated during the message-passing process. The expanding network propagation range reduces the efficiency of the latent graph inference~\cite{li2018adaptive,kazi2022differentiable}. Therefore, efforts have been made to leverage the concept of residual connections to improve models, by adding the original observed adjacency matrix (or the shallow-learned graph) to the deep-inferred adjacency matrix to rectify learning errors. Traditional residual connections involve directly summing feature matrices in Euclidean space \cite{he2016deep}. However, in non-Euclidean spaces, summation of two graphs increases invalid connections, affects predictive performance, and lacks interpretability.

This paper defines the residual connection between the original graph and the inferred graph using the Boolean product of their adjacent matrices. The Boolean product is contextualized by the essence of the graph, utilizing the original graph, the latent graph, and their corresponding fusion graph to perform triangle detection, i.e., inferring the fusion relationships between nodes from two modal graphs. For example, in a recommendation network as shown in \autoref{fig:BooleanProduct} (a), node $i$ discovers its neighbor set $\mathcal{N}_i(\mathbf{A})$ through social relationships, where $\mathbf{A}$ is a graph represents social connections, and node $j \notin \mathcal{N}_i(\mathbf{A})$. Furthermore, based on browsing records of goods, it is found that node $j$ has a neighbor set $\mathcal{N}_j(\mathbf{B})$, where $\mathbf{B}$ is a graph inferred from browsing records. When a significant intersection is observed between $\mathcal{N}_i(\mathbf{A})$ and $\mathcal{N}_j(\mathbf{B})$, the Boolean product $\mathbf{A} \diamond \mathbf{B}$ infers the presence of a connection between nodes $i$ and $j$ as shown in \autoref{fig:BooleanProduct} (b). Thus, we propose a Boolean product-based Graph Neural Network (BPGNN), which can effectively rectify observational defects in explicit graphs and update parameters in the latent graphs. Experimental results on multiple datasets and ablation studies demonstrate the advantages of the Boolean product-based graph residual connection over traditional methods.

The contributions of this paper are as follows:
\begin{itemize}
    \item Firstly, we have discovered that the Boolean product of graphs provides a more interpretable graph fusion compared to the summation of two graphs.
    \item Additionally, we have designed a Boolean product-based graph neural network, which performs message-passing via topological residual connections.
    \item Lastly, we have validated the effectiveness and robustness of the model on real-world datasets.
\end{itemize}

The rest of the paper is organized as follows. In Section 2, we review the literature relevant to our method and algorithm. We present the architecture of Boolean Product Graph Neural Networks in Section 3. The experimental results are presented in Section 4, and Section 5 concludes with some remarks.

\section{Related Works}

GNNs have emerged as crucial tools for processing structured graph data, \text{}{and achieved significant success in various domains such as social networks~\cite{guo2020deep}, bioinformatics~\cite{li2021graph} and recommendation systems~\cite{wu2022graph}. }Traditional GNNs typically operate under the assumption of a complete and accurate graph structure. However, in practical applications, one of the challenges is that the initial graph may be unavailable, incomplete, or contaminated by noise.


In scenarios where the initial graph is not given or incomplete, people have to dynamically extract structural information from the data. This requires modeling the connectivity patterns of the graph during the inference process. For instance, \citet{wang2019dynamic} proposed the EdgeConv model to handle point cloud data, which lacks explicit edge information. EdgeConv combined point cloud networks and edge convolutions to construct the graph structure. 
Similarly, ~\citet{franceschi2019learning} proposed to learn latent discrete structures by jointly learning the graph structure and GNN parameters through two levels of optimization problems. The outer-level problem aims to learn the graph structure, and the inner-level problem involves optimizing model parameters in each iteration of the outer-level problem. This approach successfully handles the application of GNNs in scenarios with incomplete or damaged graphs.


In scenarios where the observed graph data contains noise, graph fusion including multi-modal graph fusion \cite{10058089,ektefaie2023multimodal,wei2023mm,cai2022multimodal,mai2020modality} and fusing observed graphs with inference graphs \cite{chen2020iterative,sun2023self,kazi2022differentiable,de2022Latent,pan2023beyond} prove to be effective methods.

Multi-modal methods have demonstrated the potential to learn an accurate graph based on multiple observed graphs.
For example, \citet{10058089} proposed the Multimodal Fusion Graph Convolutional Network (MFGCN) model to extract spatial patterns from geographical, semantic, and functional relevance, which has been applied in accurate predictions for online taxi services.
~\citet{ektefaie2023multimodal} presented multi-modal graph AI methods that combine different inductive preferences and leverage graph processing for cross-modal dependencies. 
~\citet{wei2023mm} proposed a graph neural network model to fuse two-modal brain graphs based on diffusion tensor imaging (DTI) and functional magnetic resonance imaging (fMRI) data for the diagnosis of ASD (Autism Spectrum Disorder). 
In \cite{cai2022multimodal}, the authors consider continual graph learning by proposing the Multi-modal Structure-Evolving Continual Graph Learning (MSCGL) model, aiming to continually adapt their method to new tasks without forgetting the old ones.
\citet{pan2023beyond} utilized graph reconstruction in both feature space and structural space for clustering, effectively addressing the challenges associated with handling heterogeneous graphs.
The aforementioned multi-modal methods ignore potential modal discrepancies. To address this issue, \citet{mai2020modality} proposed an adversarial encoder-decoder-classifier to explore interactions among different modalities.

Another feasible approach is to learn a latent graph and integrate it with the initial graph. For instance, \citet{chen2020iterative} introduced the Iterative Deep Graph Learning (IDGL) model to learn an optimized graph structure. 
\citet{sun2023self} proposed a Graph Structure Learning framework guided by the Principle of Relevant Information (PRI-GSL) to identify and reveal hidden structures in a graph. \citet{kazi2022differentiable} introduced the Differentiable Graph Module (DGM), a learnable function that predicted edge probabilities in a graph, enabling the model to dynamically adjust the graph structure during the learning process. Furthermore, \citet{de2022Latent} generalized the Discrete Deep Generative Model(dDGM) for latent graph learning. 
The dDGM architecture, originally using Euclidean plane encoding of latent features, generated a latent graph based on these features. By introducing Riemannian geometry and generating a more complex embedding space, the authors enhanced the performance of the latent graph inference system.

\section{Model}
\subsection{Background}
GNNs work well for graph data $\mathcal{G}(\mathcal{V},\mathbf{A}, \mathbf{X})$, where $\mathcal{V}=\{v_1,...,v_n\}$ are the $n$ nodes with the corresponding node features $\mathbf{X}\in \mathbb{R}^{n\times d}$, and $\mathbf{A}$ is the observed adjacent relationships among nodes. However, classical GNN models assume that the underlying graph is given and fixed, which is may not satisfied in practice. 
For example, a chemical compound, merely on its molecular graph, it is hard to learn the meaningful representations of toxicity of this compound \cite{li2018adaptive}. Hence, people seek to infer a latent graph structure $\widetilde{\mathbf{A}}$ to adapt the downstream tasks with a function $F:\mathbb{R}^{n\times d}\times \mathbb{R}^{n\times n} \rightarrow \mathbb{R}^{n\times n}$ \cite{fatemi2021slaps,de2022Latent,Jianglin2023LGI}, that is 
\begin{equation*}
    \widetilde{\mathbf{A}} = F(\mathbf{X}, \mathbf{A}).
\end{equation*}
The latent graph inference takes the node features $\mathbf{X}$ and observed adjacent matrix $\mathbf{A}$ as inputs and output a latent adjacent matrix $\widetilde{\mathbf{A}}$. 
In practice, the initial observed graph is optional \cite{fatemi2021slaps,Jianglin2023LGI}. 

Currently, the graph inference function $F$ is getting more and more complicated; it could be designed as a GNN as well \cite{kazi2022differentiable}. \citet{kazi2022differentiable} proposed to learn a latent graph with a discrete Differentiable Graph Module, inferring the latent graph from aggregated features $\hat{\mathbf{X}}$,
\begin{equation*}
   \widetilde{\mathbf{A}}^{(l+1)} \leftarrow \mathbf{P}(\hat{\mathbf{X}}^{(l+1)}), 
\end{equation*}
where $\hat{\mathbf{X}}^{(l+1)}$ is the aggregated features matrix of layer $l+1$ by a GNN: $\hat{\mathbf{X}}^{(l+1)} = f_{\Theta}([\mathbf{X}^{(l)},\hat{\mathbf{X}}^{(l)}],\widetilde{\mathbf{A}}^{(l)})$.
Then another GNN module link the downstream tasks with the latent graph,
\begin{equation*}
    \mathbf{X}^{(l+1)} = \text{GNN}(\mathbf{X}^{(l)}, \widetilde{\mathbf{A}}^{(l+1)}),
\end{equation*}
where $\mathbf{X}^{(l)}$ are the latent feature of layer $l$.

As the above model go deeper and deeper, a residual connection on topology has been used in many works \cite{li2018adaptive,kazi2022differentiable,de2022Latent,lv2023robust}, 

\begin{equation*}\label{eq:originalGraphResidual}
    \mathbf{A}^{(l+1)} \leftarrow \mathbf{A} + \alpha F(\mathbf{X}^{(l)}, \mathbf{A}^{(l)}) .
\end{equation*}

\subsection{The Proposed Model}
The residual connection \cite{he2016deep} is one of the most famous techniques in deep learning, which only works in Euclidean space. 
However, the adjacent matrix lies in non-Euclidean space, making it inappropriate to perform a residual connection for graph data by simply adding two adjacency matrices.
In this work, we propose a Boolean product-based residual connection for graphs that integrates the observed adjacent matrix $\mathbf{A}$ with  $\widetilde{\mathbf{A}}^{(l+1)}$  constructed from features $\mathbf{X}^{(l)}$ as follows,
\begin{equation}\label{eq:booleanBetweenAdjacents}
    \mathbf{A}^{(l+1)} \leftarrow \mathbf{A}  \diamond \widetilde{\mathbf{A}}^{(l+1)},
\end{equation}
where $\widetilde{\mathbf{A}}^{(l+1)} =  F(\mathbf{X}^{(l)}, \mathbf{A}^{(l)})$,  and $\diamond$ is the Boolean product between two adjacency matrices, which will be introduced in next section.

\subsubsection{Boolean Matrix Multiplication}
Considering two binary matrices $\mathbf{M}, \mathbf{N} \in \{0,1\}^{n \times n}$, Boolean Matrix Multiplication \cite{booleanMM1971} between $\mathbf{M}$ and $\mathbf{N}$ is defined as following,
\begin{equation}\label{eq:BooleanMM}
    [\mathbf{M}\diamond \mathbf{N}]_{ij} = \bigvee_{k}(\mathbf{M}_{ik}\bigwedge \mathbf{N}_{kj}),
\end{equation}
where $\bigwedge$ and $\bigvee$ are Boolean \texttt{AND} and \texttt{OR} logical operators respectively.

\begin{proposition}\label{proposition:1}
    Given two graph $\mathcal{G}_1=(\mathcal{V}, \mathcal{E}_1, \mathbf{A})$, $\mathcal{G}_2=(\mathcal{V}, \mathcal{E}_2, \mathbf{B})$ that share the same node set $\mathcal{V}$, where $\mathcal{E}_1$, $\mathcal{E}_2$ are edge sets, $\mathbf{A}$ and $\mathbf{B} \in \{0,1\}^{n \times n}$ are adjacent matrices, then $\mathbf{A}\diamond \mathbf{B}$ is equivalent to  triangle detection over the graphs $\mathbf{A}$ and $\mathbf{B}$.
\end{proposition}

\autoref{fig:BooleanProduct} exemplifies the Boolean product operation applied to two graphs with adjacency matrices $\mathbf{A}$ and  $\mathbf{B}$. The nodes in the graphs are aligned. In both perspectives represented by $\mathbf{A}$ and $\mathbf{B}$, nodes $i$ and $j$ are disconnected, as indicated by $\mathbf{A}_{ij}=0$ and $\mathbf{B}_{ij}=0$. However, owing to the existence of shared neighbors  $\{1,\cdots,k\}$ between nodes $i$ and $j$, the Boolean product yields $[\mathbf{A}\diamond \mathbf{B}]_{ij}=1$. 
One special case is $\mathbf{A}\diamond \mathbf{A}$, it can be intuitively interpreted as triangle detection on $\mathcal{G}_1$.

\subsubsection{Network Architecture}
\begin{figure}[!ht]
\centering
   \includegraphics[width=0.55\linewidth]{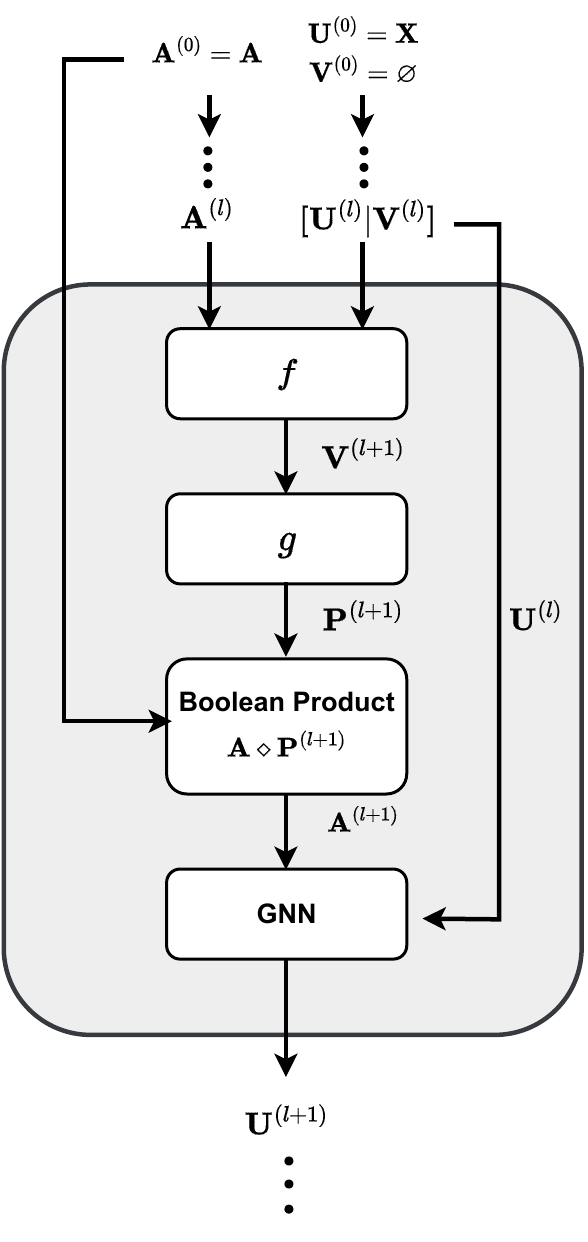}

   \caption{The basic architecture of $l$-layer of BPGNN.}
   \label{fig:Model}
\end{figure}

Figure \ref{fig:Model} illustrates the architecture of the proposed model. We decompose the latent graph inference function $F$ into a composite function of two modules, $F=f\circ g$, where $f$ is a feature aggregated function based on a graph and $g$ is a probabilistic graph learning function using the aggregated feature $\mathbf{V}^{(l+1)}$,
\begin{align}
    \mathbf{V}^{(l+1)} &= f([\mathbf{U}^{(l)}|\mathbf{V}^{(l)}], \mathbf{A}^{(l)};\mathbf{\Theta}^{(l)}),\label{eq:f4feature}\\
    \mathbf{P}^{(l+1)} &= g(\mathbf{V}^{(l+1)};\mathbf{\Phi}^{(l)}), l=0,1,...,L-1, \label{eq:g4P}
\end{align}
with $\mathbf{A}^{(0)}\leftarrow \mathbf{A}$, $\mathbf{U}^{(0)} \leftarrow \mathbf{X}$, and $\mathbf{V}^{(0)} \leftarrow \varnothing$.
This function yields a probabilistic adjacent matrix $\mathbf{P}^{(l+1)}$ as an output. Then we conduct a Boolean product between the observed adjacent matrix $\mathbf{A}$ and $\mathbf{P}^{(l+1)}$ as stated in \autoref{eq:booleanBetweenAdjacents},
\begin{equation}\label{eq:booleanFusion}
   \mathbf{A}^{(l+1)} = s(\mathbf{A}\diamond \mathbf{P}^{(l+1)}).
\end{equation}
Since $\mathbf{A}\diamond \mathbf{P}^{(l+1)}$ is a dense matrix, we further sample it into a sparse matrix using the function $s(\cdot)$. 

In~\citet{kazi2022differentiable}, $s(\cdot)$ is implemented by performing the Gumbel-Top-$k$ trick \cite{kool2019stochastic} with degree $k$. Specifically, $\mathcal{N}_i (\mathbf{A}^{(l+1)})$ is formed by the indices $j$ corresponding to the top-k largest elements from $\log(\mathbf{\tilde{p}}^{(l+1)}_i)-\log(-\log(\mathbf{q}))$, where $\mathbf{q}\in\mathbb{R}^n$ is uniformly independently distributed in the interval $[0, 1]$ and $\mathcal{N}_i (\mathbf{A}^{(l+1)})$ denotes the neighbor set of node $i$ in adjacent matrix $\mathbf{A}^{(l+1)}$.

The following module is a GNN layer, which accepts $\mathbf{A}^{(l+1)}$ and $\mathbf{U}^{(l)}$ as inputs,
\begin{equation*}
    \mathbf{U}^{(l+1)} = \text{GNN}( \mathbf{A}^{(l+1)}, \mathbf{U}^{(l)};\mathbf{W}^{(l)}),
\end{equation*}
and outputs $\mathbf{U}^{(L)}$ at the final layer $L-1$. $\mathbf{U}^{(L)}$ contains the necessary information of the downstream tasks. $\{\mathbf{\Theta}^{(l)},\mathbf{\Phi}^{(l)},\mathbf{W}^{(l)}\}$ are the parameters of layer $l=0,1,2,...,L-1$.

In practice, we implement $f$ using a GNN module. We implement $g$ following \cite{kazi2022differentiable} as $\mathbf{P}_{ij} = g(\mathbf{V}) = e^{-\gamma(\mathbf{V}_i, \mathbf{V}_j)^2 / \phi}$, where $\gamma(\mathbf{V}_i, \mathbf{V}_j)$ is a distance metric function between $\mathbf{V}_i$ and $\mathbf{V}_j$, and $\phi$ is a learnable parameter.

\subsubsection{Probabilistic Boolean product}
\autoref{eq:BooleanMM} defines the normal form of Boolean matrix product between two binary matrices. However, the matrix $\mathbf{P}$ in \autoref{eq:booleanFusion} is a continuous matrix, for example, a probabilistic matrix defined by \citet{kazi2022differentiable}.
In this section, we extend the definition in \autoref{eq:BooleanMM} to a general form, probabilistic Boolean product between a binary adjacent matrix $\mathbf{A}\in \{0,1\}^{n\times n}$ and a continuous probability matrix $\mathbf{P}^{(l+1)} \in \mathbb{R}^{n\times n}$.
Suppose $\mathbf{A}$ denotes the observed adjacent matrix, and $\mathbf{P}$ denotes the probability matrix defined in \autoref{eq:g4P}, then the probabilistic Boolean product is defined as following,
\begin{equation}\label{eq:PBMM}
    [\mathbf{A} \diamond \mathbf{P}^{(l+1)}]_{ij} = \frac{1}{\left| \mathcal{N}_i (\mathbf{A}) \right|}(\sum_{k \in \mathcal{N}_i(\mathbf{A})}\mathbf{P}_{kj}^{(l+1)}),
\end{equation}
where $\mathcal{N}_i (\mathbf{A})$ denote the neighbor set of node $i$ in adjacent matrix $\mathbf{A}$, $|\cdot|$ is the cardinality of a set.
We see that the logical \texttt{AND} and \texttt{OR} operations are replaced by product and sum respectively. 
The more interesting thing is that we find that the probabilistic Boolean product defined above becomes a message ($\mathbf{P}$) aggregation process without parameters.

We also have a symmetric version of the probabilistic Boolean product, 
\begin{equation*}\label{eq:symmetricPBMM}
     \left[\frac{\mathbf{A} \diamond \mathbf{P}+ \mathbf{P}\diamond \mathbf{A}}{2}\right]_{ij} = \frac{\sum_k \mathbf{P}_{kj}} {2\left| \mathcal{N}_i (\mathbf{A}) \right|} + \frac{\sum_k\mathbf{P}_{ik}}{2\left| \mathcal{N}_j(\mathbf{A}) \right|}.
\end{equation*}

\subsection{Loss Function}
The parameters within the latent graph inference module, i.e.,  $\mathbf{\Theta}$ and $\mathbf{\Phi}$, cannot be optimized solely through cross-entropy loss. We introduce a graph loss, as described in \citet{kazi2022differentiable}, which rewards edges contributing to correct classification and penalizes edges leading to misclassification.

Let $\hat{\mathbf{y}} = (\hat{y}_1,\cdots,\hat{y}_i)$ denotes node labels predicted by our model and the vector of truth labels is denoted as $\mathbf{y}$. The graph loss function is as follows,
\begin{equation*}
L_{\operatorname{graph}} = \sum_{i=1}^{n} \sum_{l=0}^{L-1} \sum_{j \in \mathcal{N}_i(\mathbf{A}^{(l)})} \delta(y_i, \hat{y}_i) \log \tilde{p}_{ij}^{(l+1)}(\mathbf{\Theta}^{(l)}, {\Phi}^{(l)}),
\end{equation*}
where ${\tilde{p}}_{ij}^{(l+1)} = [\mathbf{A}\diamond\mathbf{P}^{(l+1)}]_{ij}$. Additionally, $\delta\left(y_i, \hat{y}_i\right)$ denotes the reward function, indicating the difference between the average predicted accuracy and the current accuracy for node $i$. Specifically, $\delta(y_{i},\hat{y}_{i})=\mathbb{E}((a_{i}))-a_{i}$, where $a_i=1$ if $\hat{y}_i=y_i$ and 0 otherwise. 


\autoref{alg:model} represents the computational procedure of the proposed model.
\begin{algorithm}[!htbp]
   \caption{Boolean Product GNN}
   \label{alg:model}
\begin{algorithmic}
   \STATE {\bfseries Input:}  Node feature matrix $\mathbf{X}$, adjacency matrix $\mathbf{A}$
   \STATE {\bfseries Output:} Predicted node labels $\mathbf{\hat{y}}$
   \STATE {\bfseries Initialize:} $\mathbf{U}^{(0)} \leftarrow \mathbf{X}$;\, $\mathbf{A}^{(0)} \leftarrow \mathbf{A}$;\,$\mathbf{V}^{(0)} \leftarrow \mathbf{\emptyset}$
   \FOR{$l \in \{0,1,\cdots,L-1\}$}
        \STATE $\mathbf{V}^{(l+1)} \leftarrow f([\mathbf{U}^{(l)}|\mathbf{V}^{(l)}], \mathbf{A}^{(l)};\mathbf{\Theta}^{(l)})$
        \STATE $\mathbf{P}^{(l+1)} \leftarrow g(\mathbf{V}^{(l+1)};\mathbf{\Phi}^{(l)})$
        \STATE $\widetilde{\mathbf{P}}^{(l+1)} \leftarrow \mathbf{A} \diamond\mathbf{P}^{(l+1)}$
        \STATE $\mathbf{A}^{(l+1)} \leftarrow \text{Gumble-top-}k(\widetilde{\mathbf{P}}^{(l+1)})$ 
        \STATE $\mathbf{U}^{(l+1)} \leftarrow \text{GNN}( \mathbf{A}^{(l+1)}, \mathbf{U}^{(l)};\mathbf{W}^{(l)})$
    \ENDFOR 
    \STATE $\mathbf{\hat{y}} \leftarrow \text{MLP}(\mathbf{U}^{(L)})$
    \STATE Update $\{\mathbf{\Theta}^{(l)},\mathbf{\Phi}^{(l)},\mathbf{W}^{(l)}\}$ with $Loss(\mathbf{y},\mathbf{\hat{y}};\mathbf{\Theta},\mathbf{\Phi},\mathbf{W})$
\end{algorithmic}
\end{algorithm}

\section{Experiments}
In this section, we study the benefits of the BPGNN and compare it with state-of-the-art methods on node classification task. 

\subsection{Datasets and Setup}
We use four popular graph datasets: \texttt{Cora}, \texttt{PubMed}, \texttt{CiteSeer} \cite{sen2008collective}, and \texttt{Photo} \cite{shchur2018pitfalls}. The first three, \texttt{Cora}, \texttt{PubMed}, and \texttt{CiteSeer}, are citation datasets, where nodes represent scientific publications described by word vectors, and edges denote citation relationships between nodes. On the other hand, \texttt{Photo} is segments of the Amazon co-purchase graph \cite{mcauley2015image}, where nodes represent goods described by bag-of-words encoded product reviews, and edges indicate items that are frequently bought together. Please refer to \autoref{details of datasets} for detailed statistics of datasets.

In this experiment, we focus on the transductive node classification task, where all nodes are observed during training but only the train set has labels.
For our model, we employ two Boolean residual layers for \texttt{Cora} and \texttt{CiteSeer}, and a single Boolean product layer for \texttt{PubMed} and \texttt{Photo}. Additionally, 
GCN is used as aggregate function with three layers having hidden dimensions of 32, 16, and 8. Training involves the Adam optimizer with a learning rate set at $5 \times 10^{-3}$. The train/validation/test splits and all other settings follow \citet{kazi2022differentiable}. For each dataset, we execute 10 runs with different random seeds and report the mean accuracy and standard deviation.

\begin{table}[ht]
    \centering
    \caption{Summary of datasets }
    \scalebox{0.9}{
        \begin{tabular}{ccccccc}
            \toprule
            &  Cora & CiteSeer & PubMed & Photo \\
            \midrule
             \# Nodes & 2708 & 3327 & 19717 & 7650\\
             \# Edges & 5278 & 4552 & 44324 & 119081\\
             \# Features & 1433 & 3703 & 500 & 745\\
            \# Classes & 7 & 6 & 3 & 8\\
            Average Degree & 3.9 & 2.7 & 4.5 & 31.1\\
            \bottomrule
        \end{tabular}
    }
    \label{details of datasets}
\end{table}

\subsection{Baselines}
To evaluate our method, we consider some baselines 
as follows,

\begin{enumerate}
    \item MLP (Multi-layer Perceptron) : MLP neglects the graph structure.
    \item Classical GNNs.
    \begin{itemize}
        \item GCN (Graph Convolutional Network) \citep{kipf2017semi}: GCN performs a simple diffusion operation over node features;
        \item GAT (Graph Attention Network) \citep{velivckovic2018graph}: GAT refines the diffusion process by learning per-edge weights through an attention mechanism.
    \end{itemize}
    \item Latent graph inference models that only accept node features as the input.
    \begin{itemize}
        \item $k$NN-GCN: $k$NN-GCN constructs a $k$NN (sparse $k$-nearest neighbor) graph based on node feature similarities and subsequently feeds it into GCN;
        \item SLAPS \cite{fatemi2021slaps}: SLAPS provides more supervision for inferring a graph structure through self-supervision.
    \end{itemize}
    \item Latent graph inference models that accept node features  and original graph structure.
    \begin{itemize}

        \item LDS \cite{franceschi2019learning}: LDS jointly learns the graph structure and parameters of a GCN;
        \item IDGL \cite{chen2019deep} : IDGL jointly and iteratively learns graph structure and graph embedding
        \item IDGL-ANCH \cite{chen2020iterative} : IDGL-ANCH is a variant of IDGL, which reduces time complexity through anchor-based approximation \cite{liu2010large};
        \item dDGM \cite{kazi2022differentiable}: dDGM is a learnable function that predicts a sparse adjacency matrix which is optimal for downstream task.
        We use Euclidean and hyperbolic space geometries for the graph embedding space with GCN as the aggregation function, denoted as dDGM-E and dDGM-H, respectively;
        \item dDGM-EHH and dDGM-SS \cite{de2022Latent}:  dDGM-EHH and dDGM-SS incorporates Riemannian geometry into the dDGM, representing embedding spaces with a torus and a manifold of Euclidean plane and two hyperboloids.
    \end{itemize}

\end{enumerate}

\subsection{Performance}
\subsubsection{Results}
The results of the baselines and the proposed model are reported in \autoref{tab:results of models}. The results of baselines, excluding MLP, GAT, GCN and the results on the \texttt{Photo}, are sourced from the respective official reports. We can see that our model consistently outperforms all baselines on four datasets, indicating the efficacy of the proposed method. 

Firstly, we compare MLP with $k$NN-GCN and SLAPS, which rely solely on node features. It is observed that $k$NN-GCN and SLAPS perform better than MLP on \texttt{Cora} and \texttt{CiteSeer} but worse on \texttt{PubMed} and \texttt{Photo}. This emphasizes the importance of graph structure, with a lower-quality graph structure negatively impacting model performance.

Further comparison of latent graph inference models involves those utilizing only node features ($k$NN-GCN, SLAPS) and those using both node features and the original graph (LDS, IDGL, dDGM). 
It is worth noting that LDS, IDGL, and dDGM incorporate the original graph as part of their input and achieve better performance than the methods that only utilize node features. This suggests that the collaborative use of both original graph structures and node features contributes to the performance.

We explore different graph embedding spaces within dDGM models (dDGM-E, dDGM-H, dDGM-EHH, dDGM-SS). Among them, dDGM-EHH performs best on \texttt{Cora} and \texttt{CiteSeer}, while dDGM-SS performs best on \texttt{PubMed}. This indicates that the choice of embedding space impacts the similarity measurement matrix and latent graph structures. To further demonstrate the superiority of our proposed BPGNN, we compare it with dDGM-E. We observe a significant performance improvement in our model. This indicates BPGNN can help to avoid learning the wrong latent graph structures when the depth is not enough. 

\begin{table}[ht]
\centering
\caption{Results of accuracy on nodes classification task for the baselines and the proposed method. We report the mean and standard deviation (in percent) of accuracy on 10 runs for MLP, GCN, GAT and our model, as well as all for \texttt{Photo}. The others are obtained from the respective official reports. OOM indicates out of memory, OOT indicates out of time.}
\scalebox{0.85}{
\begin{tabular}{ccccc}
\toprule
Methods & Cora & CiteSeer & PubMed & Photo\\
\midrule
MLP & $62.98_{ \pm 2.624}$ & $65.06_{ \pm 3.469}$ & $85.26_{ \pm 0.633}$ &$69.60_{ \pm 3.800}$ \\
GCN & $78.74_{ \pm 1.250}$ & $67.74_{ \pm 1.723}$ & $83.60_{ \pm 1.233}$ &$92.82_{ \pm 0.653}$ \\
GAT & $80.10_{ \pm 1.672}$ & $67.74_{ \pm 1.723}$ & $82.56_{ \pm 1.436}$ &$91.80_{ \pm 1.428}$ \\
\midrule
$k$NN-GCN & $66.50_{ \pm 0.400}$ & $68.30_{ \pm 1.300}$ & $70.40_{ \pm 0.400}$ & $78.28_{ \pm 1.676}$\\

SLAPS & $74.20_{ \pm 0.500}$ & $73.10_{ \pm 1.000}$ & $74.30_{ \pm 1.400}$&$46.72_{ \pm 0.110}$ \\
\midrule
LDS & $84.08_{ \pm 0.400}$ & $75.04_{ \pm 0.400}$ &  OOT  &OOT\\
IDGL & $84.50_{ \pm 0.300}$ & $74.10_{ \pm 0.200}$ &  OOM &$90.13_{ \pm 0.200}$ \\
IDGL-ANCH & $84.40_{ \pm 0.200}$ & $72.00_{ \pm 1.000}$ & $83.00_{ \pm 0.200}$&$87.60_{ \pm 0.320}$ \\

dDGM-E & $84.60_{ \pm 0.852}$ & $74.80_{ \pm 0.924}$ & $87.60_{ \pm 0.751}$&$93.06_{ \pm 0.670}$ \\
dDGM-H& $84.40_{ \pm 1.700}$ & $74.60_{ \pm 0.763}$ & $86.60_{ \pm 0.952}$& $91.48_{ \pm 2.871}$\\
dDGM-EHH & $86.63_{ \pm 3.250}$ & $75.42_{ \pm 2.390}$ & $39.93_{ \pm 1.350}$&$--$ \\
dDGM-SS & $65.96_{ \pm 9.460}$ & $59.16_{ \pm 5.960}$ & $87.82_{ \pm 0.590}$&$--$ \\
\midrule
BPGNN  & $\bm{87.10_{ \pm 0.326}}$ & $\bm{76.34_{ \pm 0.580}}$ & $\bm{88.40_{ \pm 0.770}}$ &$\bm{93.40_{ \pm 0.604}}$\\
\bottomrule
\end{tabular}
}
\label{tab:results of models}
\end{table}

\subsubsection{Robustness}
\begin{figure}[ht]
    \centering
    \subfloat[Cora]{
    \begin{minipage}[t]{1 \columnwidth}
        \centering     \includegraphics[width=0.95\columnwidth]{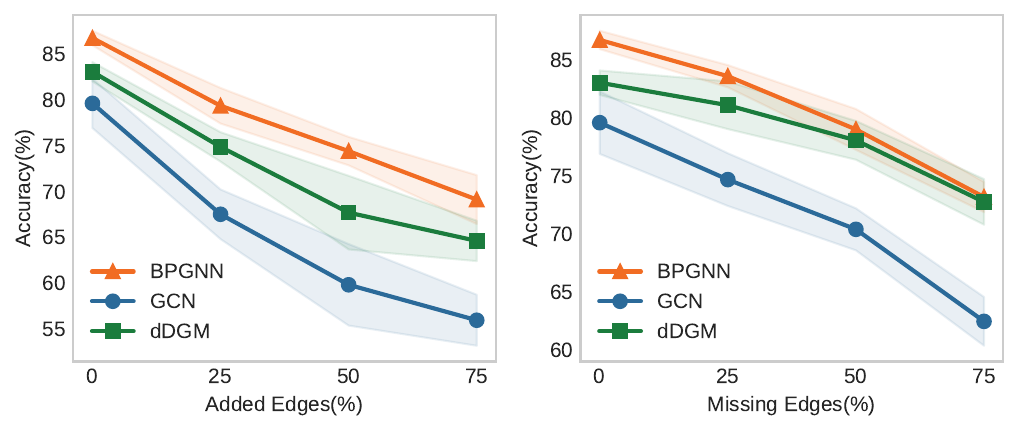}
    \end{minipage}} \\
    \subfloat[CiteSeer]{
    \begin{minipage}[t]{1\columnwidth}
        \centering
\includegraphics[width=0.95\columnwidth]{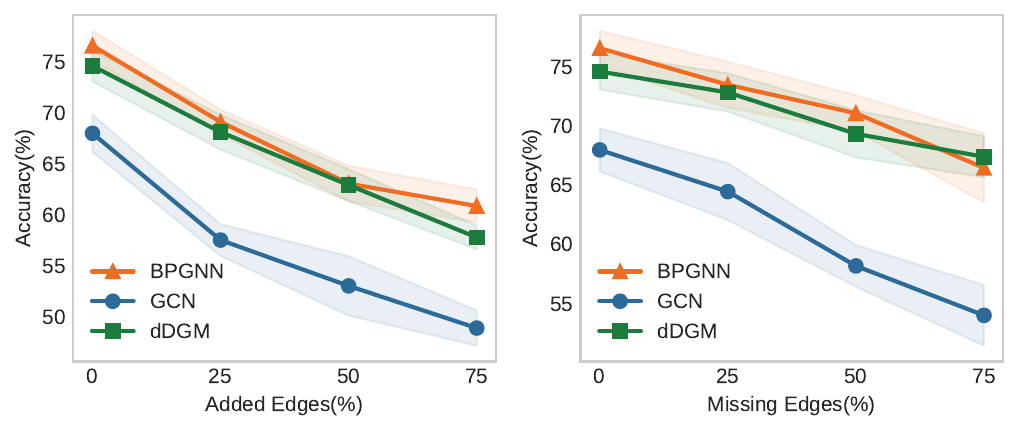}
    \end{minipage}}
   \caption{Test accuracy ($\pm$ standard deviation ) in percent for the edge addition and deletion on Cora and CiteSeer.}
   \label{fig:noise}
\end{figure}

To assess the robustness of BPGNN, we construct graphs with random edges deletion and addition on \texttt{Cora} and \texttt{CiteSeer} as shown in \autoref{fig:noise} (a) and (b) respectively. Specifically,  we add or remove edges at ratios 25\%, 50\%, and 75\% to the existing edges for each dataset. Compared to GCN and dDGM, BPGNN achieves better results in both scenarios. Especially in scenarios involving edge addition, BPGNN consistently demonstrates outstanding performance across all ratios, showcasing its robust ability to eliminate noise from the graph.

\subsubsection{Homophily}
\begin{figure}[h!t]
  \centering
\includegraphics[width=0.8\linewidth]{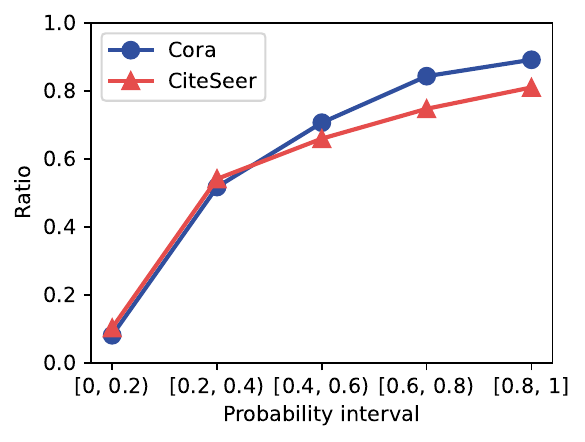}
   \caption{The ratio of two nodes in the test set sharing the same label in different probability interval.}
   \label{fig:odds}
\end{figure}

To evaluate the quality of learned graph structure concerning homophily, 
we compute the ratio of the number of nodes pairs with the same label to the total number of nodes pairs in different probability intervals, using the test set of \texttt{Cora} and \texttt{CiteSeer}, as shown in \autoref{fig:odds}. We can see that the higher the probability of a connection between two nodes by \autoref{eq:g4P}, the more likely it is that their labels are consistent. This indicates that BPGNN effectively captures and reinforces homophilic relationships in the graph structure.

\subsection{Ablation Study}
In this section, we conduct extensive ablation experiments to demonstrate the performance of the key components of the proposed model.
\subsubsection{Number of Boolean Product Layers}
In this experiment, we investigate the impact of using the Boolean product module rather than just performing GNNs on input graph and explore whether there is any advantage in stacking multiple Boolean product layers. The results, detailed in \autoref{number of layers}, show the performance concerning the number of Boolean product layers, where a value of ``0'' indicates the absence of the Boolean product module, and GCN is just used.

It is observed that using Boolean product layers alone is sufficient to enhance classification accuracy. Cora and CiteSeer benefit from a 2-layer Boolean product, whereas for PubMed, the optimal number of layers is one. The potential reason could be that Cora and CiteSeer have a smaller number of nodes, and the original graph is relatively simple. They require multiple Boolean product layers to infer the latent graph structures.  In contrast, PubMed's more complex original graph structure, containing richer information, allows a single Boolean product layer to infer a high-performing latent graph structure.

\begin{table}[ht]
\centering
\caption{ Results of Boolean Residual model with a different number of Boolean product layers. We report the mean and standard deviation (in percent) of the accuracy on 10 runs.}
\scalebox{0.9}{
\begin{tabular}{cccc}
\hline Layers & Cora & CiteSeer & PubMed \\
\hline 
0 & $78.74_{ \pm 1.25}$ & $67.74_{ \pm 1.723}$ & $83.60_{ \pm 1.233}$ \\
1 & $86.36_{ \pm 0.920}$ & $75.92_{ \pm 1.200}$ & $\bm{88.40_{ \pm 0.770}}$ \\
2 & $\bm{87.10_{ \pm 0.326}}$ & $\bm{76.34_{ \pm 0.580}}$ & $87.12_{ \pm 0.859}$ \\
3& $86.82_{ \pm 0.817}$ & $75.54_{ \pm 0.925}$ & $86.72_{ \pm 0.531}$ \\
\hline
\end{tabular}
}
\label{number of layers}
\end{table}

\subsubsection{Aggregate function}
In our analysis of aggregate functions, we consider three options: GCN \cite{kipf2017semi}, GAT \cite{veli2018graph}, and EdgeConv \cite{wang2019dynamic}. The results, as depicted in \autoref{diffusion functions}, show comparable performance between GCN and GAT, with no significant difference. However, the adoption of EdgeConv notably reduces accuracy, particularly on small datasets.

This observation can be explained by the fact that a latent graph structure obtained through Boolean product module already incorporates the information about nodes interactions. Therefore, GAT does not significantly contribute to performance improvement. Additionally, during the training process of our model, each sampled latent graph varies, making it challenging for the loss involving edge attributes to converge. Consequently, the use of EdgeConv leads a decrease in accuracy.

\begin{table}[ht]
\centering
\caption{Results of our model with different aggregate fuctions. We report the mean and standard deviation (in percent) of the accuracy on 10 runs.}
\scalebox{0.9}{
\begin{tabular}{cccc}
\hline $\text{Aggre}(\cdot)$ & Cora & CiteSeer & PubMed \\
\hline GCN & $87.10_{ \pm 0.326}$ & $\mathbf{76.34_{ \pm 0.580}}$ & $\mathbf{88.40_{ \pm 0.770}}$ \\
GAT & $\mathbf{87.30_{ \pm 1.020}}$ & $76.22_{ \pm 1.130}$ & $88.20_{ \pm 0.704}$ \\
EdgeConv & $55.70_{ \pm 5.100}$ & $52.44_{ \pm 2.890}$ & $86.08_{ \pm 0.900}$ \\
\hline
\end{tabular}
}
\label{diffusion functions}

\end{table}

\subsection{Time Complexity Analysis}
In the Boolean product module, Equation \ref{eq:PBMM} involves the computation of the matrix product between between a binary adjacent matrix $\mathbf{A}\in \{0,1\}^{n\times n}$ and a continuous probability matrix $\mathbf{P}\in \mathbb{R}^{n\times n}$, where $n$ denotes the number of nodes. Direct matrix multiplication has a computational complexity of $\mathcal{O}(n^3)$. However, considering the sparsity of matrix $\mathbf{A}$, employing sparse matrix multiplication reduces the complexity to $\mathcal{O}(\alpha n^2)$, where $\alpha$ denotes the sparsity level of $\mathbf{A}$. Here, $\alpha$ is defined as the ratio of the number of non-zero elements to the total number of elements. \autoref{details of datasets} presents the average degree of datasets, indicating that $\alpha \ll 1$, making matrix multiplication more computationally efficient.

\begin{figure}[t]
  \centering
   \includegraphics[width=0.85\linewidth]{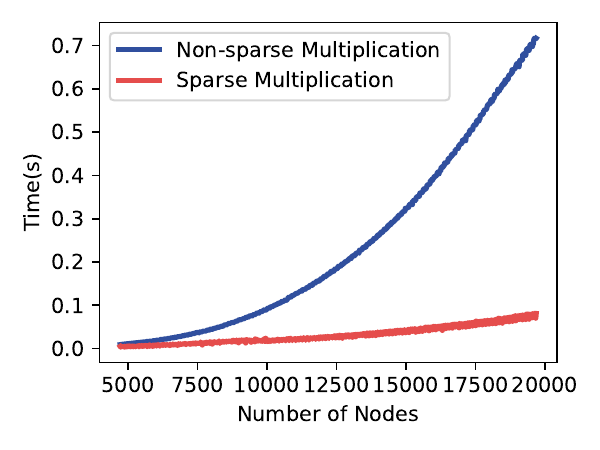}
   \caption{Time requirements for matrix multiplication under both sparsity and non-sparsity computed using the \texttt{torch.sparse.mm} and \texttt{torch.matmul} functions in PyTorch, respectively.}
   \label{fig:SpeedOfBoolean}
\end{figure}

In Figure \ref{fig:SpeedOfBoolean}, we illustrate the computational time required for Equation \ref{eq:PBMM} on input number of nodes $n$ under both sparse and non-sparse conditions. We employed the \texttt{torch.matmul} and \texttt{torch.sparse.mm} from the PyTorch package to implement matrix multiplication for both scenarios. The adjacency matrix of PubMed serves as $\mathbf{A}$, and we derive $\mathbf{P}$ using its node features. Subsequently, by randomly dropping nodes, we obtain the computation time for matrix multiplication with different numbers of nodes. We can see that sparse matrix multiplication results in faster computation, particularly on larger graphs $n \gtrapprox 5000$, overcoming the cubic complexity.

\section{Conclusion}
The graph structure is very important for GNNs. Many studies on latent graph inference have confirmed that graph structure noise is prevalent in popular graph datasets. The noise tends to  be amplified during the message-passing process, impacting the performance of GNNs. To address this issue,  this paper introduces Boolean residual connection to improve performance of the dDGM model. Our experimental results on three widely-used graph datasets for nodes classification task demonstrate the superior performance of our proposed model.

In this study, we adopt the Boolean product as the residual connection method. Besides that, it would be interesting to explore other residual connection methods between the graph structures. We hope our work can inspire more research into latent graph inference, able to infer graphs that are closer to the true underlying graph.

\bibliography{references}


\end{document}